\documentclass[runningheads]{llncs}
\usepackage[T1]{fontenc}
\usepackage{graphicx}
\usepackage{booktabs}
\usepackage{hyperref}
\usepackage{xcolor}
\usepackage{subfig}

\usepackage{tikz}
\usetikzlibrary{shapes.misc,decorations.pathreplacing}
\usetikzlibrary{patterns}
\usepackage{xurl}

\begin{document}
\title{Cleaning Robots in Public Spaces: \\ A Survey and Proposal for Benchmarking \\ Based on Stakeholders Interviews}
\titlerunning{Cleaning Robots in Public Spaces}
%
\author{Raphael Memmesheimer\inst{1}
\and Martina Overbeck\inst{2}
\and Bjoern Kral\inst{3}
\and Lea Steffen \inst{4}
\and Sven Behnke\inst{1}
\and Martin Gersch\inst{3}
\and Arne Roennau\inst{4}
}
\authorrunning{R. Memmesheimer et al.}
%
\institute{Autonomous Intelligent Systems, Computer Science, University of Bonn, Germany
    \and FZI Forschungszentrum Informatik, Germany
    \and Business Administration, Free University of Berlin, Germany
    \and Intelligent Machines, Karlsruhe Institute of Technology, Germany
}
\maketitle              
\begin{abstract}
Autonomous cleaning robots for public spaces have potential for addressing current societal challenges, such as labor shortages and cleanliness in public spaces. 
Other application domains like autonomous driving, bin picking, or search and rescue have shown that benchmarking platforms and approaches in competitive settings can advance their respective research fields, resulting in more applicable systems under real-world conditions. 
For this paper, we analyzed seven semi-structured, qualitative stakeholder interviews about outdoor cleaning, identified current needs as well as limitations, and considered those results for the development of a benchmarking scenario based on the previous observations. 
\keywords{cleaning robots  \and expert interviews \and robotic competitions \and benchmarking.}

\end{abstract}
\section{Introduction}
Automation has become an integral part of our daily lives and has transformed various industries and sectors. One such sector that has seen notable advances is the automated cleaning industry. 
Although considerable research has been dedicated to the development of private autonomous indoor cleaning robots~\cite{DBLP:journals/ijsr/FinkBKD13,DBLP:conf/icra/MohanRC14,DBLP:journals/arobots/FioriniP00}, the exploration of autonomous cleaning robots for public spaces has not yet received the same analysis. Automated public cleaning robots have the potential to address societal challenges, such as maintaining cleanliness and offering a promising solution to labor shortages in the public cleaning sector.

We researched public sector requirements via stakeholder interviews and an analysis of technological requirements for cleaning robots. Based on both and further motivated by the positive impact of robotic competitions, we propose a benchmarking scenario to foster the development of cleaning robots for public spaces.

\noindent The contributions of this paper are as follows:
\begin{itemize}
    \item \textit{Expert Interviews}: Seven expert interviews were conducted to gain a more profound understanding of the requirements and challenges in communal cleaning of public areas. These interviews provided valuable insights into stakeholders' present requirements and expectations in this domain. 
    \item \textit{Identification of Limitations}: This research identified several limitations of current hardware and software approaches to autonomous cleaning of public spaces. 
    \item \textit{Benchmark Scenario}: Based on our expert interviews and an analysis of current limitations, we developed a benchmarking scenario aimed at fostering advancements in autonomous cleaning in public spaces.
\end{itemize}


\section{Related Work}

In the following, we give a brief overview of available cleaning robot solutions and robotic competitions. 


\subsection{Robotic and Automated Cleaning}

Vacuum cleaning robots came onto the market in the early 2000s~\cite{DBLP:journals/arobots/FioriniP00,Prassler2000} and paved the way for the development and introduction of other cleaning robots for different applications and environments. Husqvarna's robotic lawnmower Solarmower, which appeared in 1996, is considered the role model for the first robotic vacuum cleaners~\cite{BBC2003}. The first robot vacuum cleaner was the Trilobite from Electrolux, which was commercially available from 2001~\cite{Electrolux2006}. 

In 2002, iRobot~\cite{Wang2019} introduced the now well-known Roomba robot vacuum cleaner. In 2012, it had sold more than 8 million units~\cite{BBC2003}. The robot was able to recognize and avoid obstacles. However, this system was limited in corner cleaning capabilities. 

In the 2010s, there were many advancements and, consequently, a rapidly increasing demand for domestic robots -- mainly due to advanced sensing and navigation technologies~\cite{BBC2003}. This led to improved autonomy and flexibility of cleaning robots. Robot-based cleaning systems that were able to move autonomously through complex environments and perform various cleaning tasks became more common~\cite{Prassler2000}. Compared to today's systems, the systems of the time suffered from poor navigability, complexity, frequent failures, and low cleaning performance. In addition, the products were still large and bulky~\cite{Gausium2023}.

Since the 2020s, cleaning robots have been increasingly found in commercial scenarios, such as office buildings, shopping malls, airports, hotels, and healthcare facilities. Cleaning tasks they offer include dusting floors, vacuuming, washing windows, and more. Technology is constantly evolving, and the development of artificial intelligence (AI) and sensor technology continues to drive the performance and adaptability of these robots. Current technologies are characterized primarily by achievements in the following two areas~\cite{Gausium2023}: 

First, the cleaning performance has increased, e.g. by advanced brush design. This enables cleaning along edges and in tight corners. The battery capacity has also been increased. With a single charge and a service life of over 1,000 charging cycles, several hours of scrubbing or dusting can now be performed.

Second of all, autonomous navigation has been improved with new sensor technology and the combination of distance and collision sensors~\cite{TechHive2018}, thus enabling the replacement of inefficient pathfinding using random walks with superior algorithms such as SLAM (Neato Robotics, Samsung Hauzen RE70V, Dyson 360 Eye), fuzzy logic methods (Samsung POWERbot), and the A* algorithm (LG Hom-Bot)~\cite{Edwards2018,Raguraman2009}. 



\begin{figure}
  \centering
  \subfloat[Adlatus CR700 \cite{Adlatus}]{\includegraphics[width=.3\linewidth]{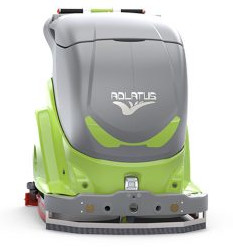}}\quad 
  \subfloat[Angsa \cite{Angsa}]{\includegraphics[width=0.3\textwidth]{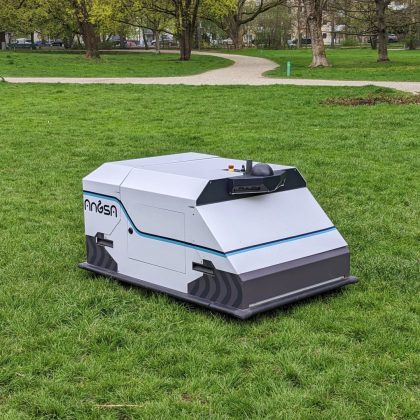}}\quad %
  \subfloat[VIGGO S100-N \cite{Viggo}]{\includegraphics[width=0.3\textwidth]{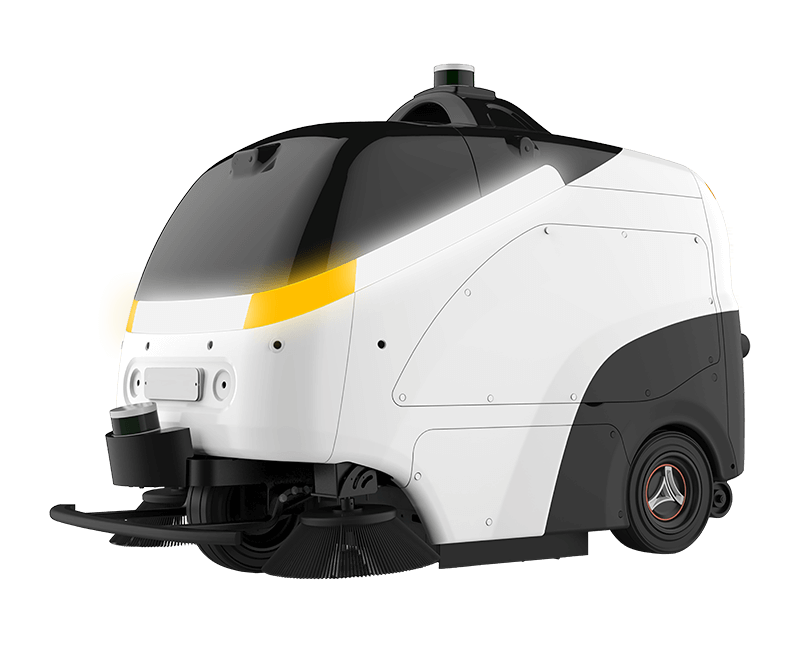}} 
  \caption{Cleaning robot platforms.}
  \label{fig:cleaning_robots}
\end{figure}

The state of the art in commercial, automated cleaning technologies already includes a wide range of proven and established systems. 
The most commonly required cleaning tasks are classified into six robotic motions: wiping, sweeping, scrubbing, vacuuming, washing and tidying up~\cite{DBLP:journals/arobots/FioriniP00,Prassler2000}. An excerpt of existing cleaning robots for public spaces are depicted in Figure~\ref{fig:cleaning_robots}.

The integration of simple AI methods, robotics, and the Internet of Things (IoT) into cleaning technologies has led to a significant improvement in efficiency, quality, sustainability, and user-friendliness. Systems have already been successfully tested and introduced in the following areas: Firstly, autonomous robots can use sensors and 2D path planning~\cite{Edwards2018} to recognize their surroundings and avoid obstacles. They can be used in various environments, including offices, shopping centers, airports, and warehouses. Such systems are capable of mopping floors, vacuuming carpets, cleaning windows, and other routine tasks. Drones are increasingly being used for outdoor cleaning, especially for hard-to-reach outdoor areas such as high-rise facades or solar panels. These drones are equipped with special cleaning agents and brushes and can work autonomously or remotely. A study specifically on robot-based cleaning of solar installations was presented in~\cite{Hashim2019}.





Based on this assessment of the current status, it is possible to identify gaps and weaknesses in existing work on robotic cleaning systems. The combination of different sensors, such as cameras, LiDAR, and infrared sensors for detecting walls, steps, and stairs or encoders for measuring wheel rotation, enables advanced navigation with high precision. 

Cleaning systems are currently unable to adapt flexibly to changing environments and only take 2D surfaces into account. For example, future robotic cleaning systems should be able to cope well with changing conditions in a natural environment. This includes darkness, reflections, extreme solar radiation, and heat, as well as heavy rain or wind. On the other hand, an extension to 3D planning would be helpful and would allow the observation of more complex environments.

In addition, cleaning should only take place where it is necessary. This could lead to reduction in water, energy, and chemical consumption as well as a multiple increase in overall efficiency. There are already initial systems that can identify soiling~\cite{Gausium2023,Bormann2020}. Nevertheless, there is still great potential for innovative developments regarding the identification of contaminated areas to enable the cleaning of selected areas. 

Visual perception methods based on recent advances of Convolutional Neural Networks (CNNs) can be applied to identify dirt and therefore focus the cleaning process. 
The design of many systems also poses a problem for mobility. The smaller the size, the greater the ability to navigate in narrow aisles, corridors, and other confined areas. However, this is in stark contrast to battery life and the associated performance, which tends to require larger designs. 

Finally, automated cleaning in public areas requires a positive user experience, as this is the basis for technology acceptance. A distinction is made here between interaction with users and observers who do not operate the device directly. 


\begin{table}
    \caption{Competition overview.}
    \resizebox{\columnwidth}{!}{%
    \begin{tabular}{l|l|l|r}
        \toprule
        \textbf{Competition} & \textbf{Focus} & \textbf{Country} & \textbf{Year} \\ \midrule
        RoboCup~\cite{pellenz2016novel,RoboCupHomeRuleboook,robocupsoccer}  & Soccer, Home, Rescue, Logistics, Junior, ... & Int.  & since 1996  \\
        DARPA Challenges~\cite{thrun2006stanley,DARPASubTRulebook} &  Military & USA & since 2004\\
        European Robotics League & Home, Rescue, Work & Europe & 2016-2023\\
        MBZIRC~\cite{MBZIRC} & Military & UAE & 2017, 2019, 2023 \\
        ANA Avatar XPRIZE~\cite{AVATARRulebook} & Telepresence & USA & 2018–2022\\
        Amazon Robotics Challenge~\cite{AmazonPickingChallengeRulebook}& Industrial / Logistics & Int. & 2015–2017 \\
        METRICS~\cite{metricsproject,MetricsRulebook} & Domestic, Industrial, Agriculture & Europe & 2020–2023 \\
        World Robot Summit~\cite{WRS,kimura2017competition,okada2019competitions,yokokohji2019assembly} & Service, Industrial, Rescue, Junior & Japan & 2018, 2021\\
        \bottomrule
    \end{tabular}
    }
\end{table}
\subsection{Robot Competitions}

Robotic competitions~\cite{robocupsoccer,GerndtSBSB:RAM15,roa2021mobile,AmazonPickingChallengeRulebook} serve as exciting platforms for researchers, engineers, and students to showcase their designs, push the boundaries of technology, and test the skills of robots against real-world challenges. In this section, we introduce various robotics competitions taking place around the globe.

\subsubsection{General Robot Competitions}

\emph{RoboCup:} International competition focusing on autonomous robots, with various applications and events, such as soccer~\cite{robocupsoccer,GerndtSBSB:RAM15}, service robots~\cite{RoboCupHomeRuleboook}, and rescue robots~\cite{pellenz2016novel}. Annual, worldwide, with more than 2000 participants. 

\emph{DARPA Challenges:} US-based competitions focusing on disaster scenarios and advanced humanoid robots and exploration of external environments. Notable events include the DARPA Grand Challenge 2006~\cite{thrun2006stanley}, the DARPA Robotics Challenge 2015, and the DARPA Subterranean Challenge 2021~\cite{DARPASubTRulebook}.

\emph{ANA Avatar XPRIZE:} US-based competition focused on developing an avatar system allowing human presence to be represented by a remotely controlled robot, with a focus on teleoperation, haptics, and interaction~\cite{AVATARRulebook,BehnkeALRAM23,hauser2024analysisSORO}. 

\emph{Amazon Picking/Robotics Challenge:} Challenge to develop robotic systems capable of autonomously picking items in a warehouse environment, with specified objects and warehouse layout~\cite{AmazonPickingChallengeRulebook}.

\emph{Metrics:} EU-funded project focusing on the development of robots in various fields, including healthcare, agile production, inspection, and agricultural robotics~\cite{metricsproject,MetricsRulebook}. 


\subsubsection{Robot Competitions with Cleaning Aspects}

The aforementioned competitions focus on various aspects of robotic automation, ranging from autonomous driving, rescue, space, domestic, industrial manufacturing and warehouse scenarios. We now introduce competitions and challenges with a cleaning aspect.

\emph{Deutsche Bahn: Automated Cleaning Challenge (2018, Germany)} Deutsche Bahn organized a competition in which robots were to be developed that could autonomously clean train stations. The winning team was awarded a two-year contract with Deutsche Bahn to further develop station cleaning. The competition was aimed at innovative companies and start-ups with a focus on automated cleaning. A rule book or a detailed description of the competition are unfortunately no longer available online~\cite{DBCleaningChallenge2023}.

\emph{World Robot Summit (2018, 2020, Japan)} The World Robot Summit (WRS)~\cite{WRS} is an international robotics competition 
with challenges ranging targeting rescue~\cite{kimura2017competition}, 
service~\cite{okada2019competitions}, 
industrial~\cite{yokokohji2019assembly} and special challenges
for juniors. In the Future Convenience Store Challenge (FCSC), robots competed against each other in a supermarket. One task in the FCSC was to clean a customer toilet. The task was semi-automatically scored by calculating a cleaning rate from before and after images under UV light.

\emph{UV Robot Design Contest (2021, online)} was a design competition for robot designs that to eliminate COVID-19 bacteria using UV radiation.

\section{Stakeholders on Cleaning Robots in Public Spaces}

To extract meaningful comparison criteria for cleaning robotics in public spaces, stakeholders' opinions were sought through interviews. A semi-structured expert interview approach, according to ~\cite{adams2015conducting} and ~\cite{kallio2016systematic}, was chosen to frame this analysis of requirements for robotic cleaning systems. This allows for adaptability and in-depth exploration of various topics based on the responses and information provided by the experts. An interview guide was created and divided into seven thematic areas addressing economic and technical needs: cleaning tasks and current challenges, the potential for automation through robotic cleaning systems, acceptance and personnel issues, purchasing behavior, potential opportunities and challenges, market development, and trends. The original and English translation of the questionnaire is made available\footnote{\url{https://www.roboter-im-alltag.org/aktuelles/publikationen/}}.
In total, 82 stakeholders from municipal and urban cleaning service providers, as well as facility management sectors, were contacted, after a conducting stakeholders research. The interviewed persons were mainly managers with a strategic focus but also in contact with executive employees in the city cleaning department. In the end, 8 interviews, each lasting approximately 42 minutes, were conducted. They revealed a keen interest in future projects within cleaning robotics for both indoor and outdoor cleaning. Only 7 interviews related to public outdoor cleaning are included here.

\begin{table}[]
\centering
\caption{Areas of application (A1-A14) for public space cleaning robots as mentioned from interviews with stakeholders (Mentions in \%).}
\label{tab:area-of-application}
\vspace*{1ex}
\resizebox{\columnwidth}{!}{%
\begin{tabular}{llr}
\toprule
\textbf{No.} & \textbf{Areas of Application (Tasks)}                                           & \textbf{M. in \%} \\ \midrule
A1           & Collecting small, possibly health-hazardous waste of different consistencies    & 100\%                   \\ 
A2           & Reaching places that are difficult to access                                    & 100\%                   \\ 
A3           & Vacuuming                                                                       & 100\%                   \\ 
A4           & Sensory detection/ recognition of objects and people, as well as their quantity & 86\%                    \\ 
A5           & Sweeping of sidewalks, squares, surfaces                                        & 86\%                    \\ 
A6           & Leaf removal                                                                    & 71\%                    \\ 
A7           & Raking/hooking for weed removal on water-bound paths/areas                      & 71\%                    \\ 
A8           & Cleaning seams on/in sidewalks or cobblestones                                  & 57\%                    \\ 
A9           & Transporting heavy objects or collected waste                                   & 57\%                    \\ 
A10          & Emptying stationary waste garbage cans                                          & 57\%                    \\ 
A11          & Snow clearing/ gritting                                                         & 57\%                    \\ 
A12          & Removing chewing gum or graffiti from surfaces                                  & 43\%                    \\ 
A13          & Mopping of sidewalks, squares, surfaces                                         & 43\%                    \\ 
A14          & Lifting heavy objects such as maintenance hole covers                           & 29\%    \\   
\bottomrule
\end{tabular}%
}
\end{table}

The evaluation of expert interviews regarding stakeholder requirements in the context of robotic cleaning systems revealed relevant insights into application areas and potentials for robotic cleaning. All interviewees highlighted tasks (Table~\ref{tab:area-of-application}) such as collecting or vacuuming small, possibly health-hazardous waste of different consistencies (e.g., crown caps, cigarettes, broken glass, candy wrappers, but also syringes, feces, condoms) as well as reaching places that are difficult to access (e.g., under cars or park benches, near water edges, tree grates and windows, or in corners). Robotic systems were, in the majority of cases, also identified as a potential asset for the sensory detection and recognition of objects and people, as well as their quantity, as well as vacuuming and sweeping sidewalks, squares, and surfaces.

The organizational context and personnel aspects also played a central role in the analysis of the economic needs. The main objective (see Table~\ref{tab:objectives}) as to why robots were interesting for outdoor cleaning tasks was the improvement of cleaning quality. That was, nonetheless, closely followed by the mention of the need to clean areas for which the cleaning staff would generally have no capacity, as well as the reduction of downtime caused by both sickness and a shortage of personnel. 

\begin{table}[]
\centering
\caption{The most relevant objectives (O1-O7) mentioned by the stakeholders in the semi-structured qualitative interviews (Mentions in \%). 
}
\label{tab:objectives}
\vspace*{1ex}
\resizebox{\columnwidth}{!}{%
\begin{tabular}{llr}
\toprule
\textbf{No.} & \textbf{Objectives}                                                            & \textbf{M. in \%} \\ \midrule
O1           & Improvements in cleaning quality                                               & 86\%                    \\ 
O2           & Cleaning of areas for which the cleaning staff would normally have no capacity & 71\%                    \\ 
O3           & Reduction of downtime                                                          & 71\%                    \\ 
O4           & Improvement in cleaning speed                                                  & 57\%                    \\ 
O5           & Long-term savings and cost benefits                                            & 43\%                    \\ 
O6           & Standardized execution of cleaning with consistent results                     & 43\%                    \\ 
O7           & Reduction of stress (physical, health) for cleaning staff                      & 43\%  \\
\bottomrule
\end{tabular}%
}
\end{table}

The interviews also tackled the question of what boundary conditions robots would have to face (Table~\ref{tab:boundary-conditions}) to fulfill both their task (e.g., picking up small objects) and the objective (e.g., increasing cleaning quality). Within the most frequently mentioned conditions were having to deal with strongly deviating, heterogeneous substrates, changing weather and lighting conditions, and the varying availability of storage and loading sites, as well as the necessity to be aware of the welfare of nature, animals, and children. Most of all, however, mentioned as a boundary condition for the use of cleaning robots in public outdoor spaces was the role of (non)acceptance of the robots that influences the interaction of both workers and passersby and has the interviewees worried about vandalism. 

\begin{table}[]
\centering
\caption{The boundary conditions for task fulfillment (B1-B15) for cleaning robots were identified from interviews with stakeholders.
}
\label{tab:boundary-conditions}
\vspace*{1ex}
\resizebox{\columnwidth}{!}{%
\begin{tabular}{llr}

\toprule
\textbf{No.} & \textbf{Boundary Conditions for Task Fulfillment}                  & \textbf{Mentions in \%} \\ 
\midrule
B1           & Dealing with (non-)acceptance, interactions, vandalism             & 100\%                   \\ 
B2           & Dealing with strongly deviating, heterogeneous substrates/surfaces & 86\%                    \\ 
B3           & Dealing with changing weather and lighting conditions              & 86\%                    \\ 
B4           & Environmental protection/ animal welfare/ child protection         & 86\%                    \\ 
B5           & Dealing with different storage and loading availability            & 86\%                    \\ 
B6           & Dealing with unmapped, unknown areas                               & 57\%                    \\ 
B7           & Coping with stairs/ steps                                          & 57\%                    \\ 
B8           & Sufficient storage capacity of the robot itself                    & 57\%                    \\ 
B9           & Dealing with grass of different heights                            & 43\%                    \\ 
B10          & Removing trampled small items of waste                             & 43\%                    \\ 
B11          & Dealing with noise/volume restrictions                             & 43\%                    \\ 
B12          & Coping with inclines                                               & 29\%                    \\ 
B13          & Achieving visual improvement through cleaning                      & 29\%                    \\ 
B14          & Vacuuming without removing or damaging the surface                 & 29\%                    \\ 
B15          & Dealing with unknown objects                                       & 14\%                   
\\
\bottomrule
\end{tabular}%
}
\end{table}

\section{Benchmarking}

In this section, based on the stakeholder interviews, we derive required basic robotic technologies and hardware requirements. We further propose a benchmarking scenario that combines the technological requirements and the mentioned applications by the stakeholders.

\subsection{Technologies and Hardware Requirements for Urban Cleaning}
Following the state of the art and the stakeholder interviews, the basic technologies required in the context of urban cleaning can be identified:

\emph{Perception} is required for environmental perception. Cameras and sensors are used to record the robot's surroundings, recognize obstacles and humans, assess the cleaning requirements, and track the progress of the cleaning. The sensor technology to be used depends on the task. A camera is used to detect contamination, a LiDAR is used to create a map of the environment and detect obstacles, and a camera or LiDAR is used to detect people. It is conceivable that active lighting or cameras in the non-visible range could be used to improve perception. Perception can often be significantly improved by fusing multiple sensors.

\emph{Manipulation} of objects is task-dependent. For example, dirt can be removed by manipulation using a suction device or a sponge. Picking up garbage can be implemented by manipulation using a gripper. Manipulation is strongly based on the results of perception. If these are deficient, manipulation is usually not possible. Manipulation tasks involving variable objects increase the complexity of the robot and usually require robot arms with several degrees of freedom. There are certain dependencies between the individual basic technologies. 

\emph{Cognition} is necessary for decision-making. The robot must be able to understand the environment and make decisions based on this. There is, therefore, a direct dependency on perception because of which decisions are made. State machines are a simple way of making decisions. However, more complex methods such as behavior trees~\cite{Heppner2019} can also be used here. 

\emph{Navigation} enables the robot to move around in an environment. This includes localization and path planning. Localization can be performed using a LiDAR, movement estimates from odometry (e.g., from wheel rotation), and path planning using a map. The necessary movement commands are transmitted to the robot via a drive and should occur in a closed control loop. There are established localization and path planning methods, which can usually be used on a robot but require initial integration effort. 

\emph{Interaction} is necessary for operating the robot. This includes operation via user interfaces, speech recognition, and gesture recognition. People are recognized by a camera or a LiDAR, speech is recognized by a microphone, and gestures are recognized by a camera or a LiDAR. Interaction with people is, therefore, closely linked to sensor-based perception.

Manipulation, cognition, navigation and interaction are all based on perception. Inadequate perception, therefore, usually has major consequences. Depending on the task, there are also dependencies between the other basic technologies and navigation.

\subsection{Benchmarking Scenario: Park Cleaning}

Park pathway cleaning is a well benchmarkable task that unifies the criteria stated by the stakeholder interviews and the identified basic technologies, and therefore serves as an interesting benchmarking scenario of cleaning robots for public spaces.

\begin{figure}
    \centering
    \includegraphics[width=0.24\linewidth]{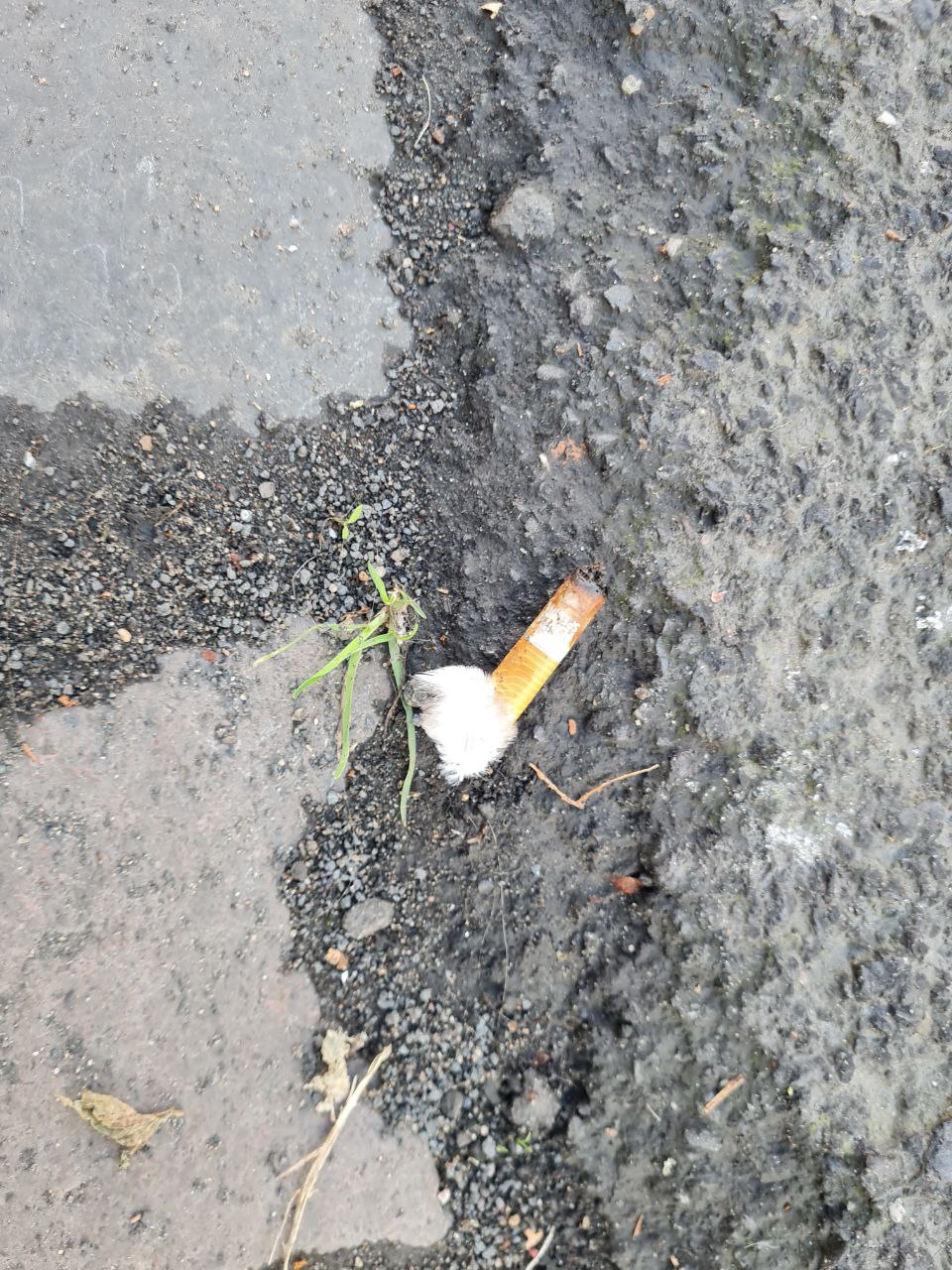} 
    \includegraphics[width=0.24\linewidth]{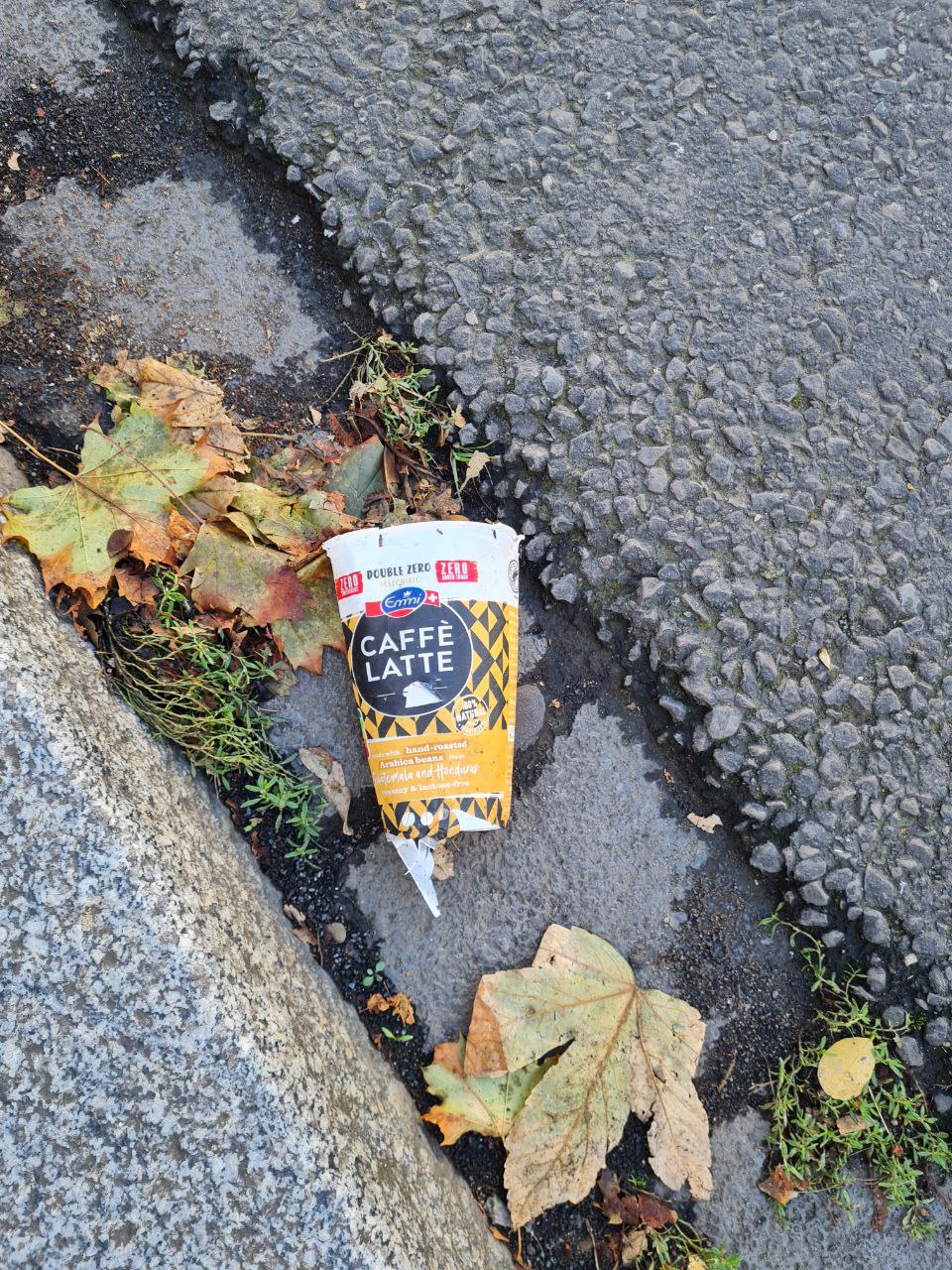} 
    \includegraphics[width=0.24\linewidth]{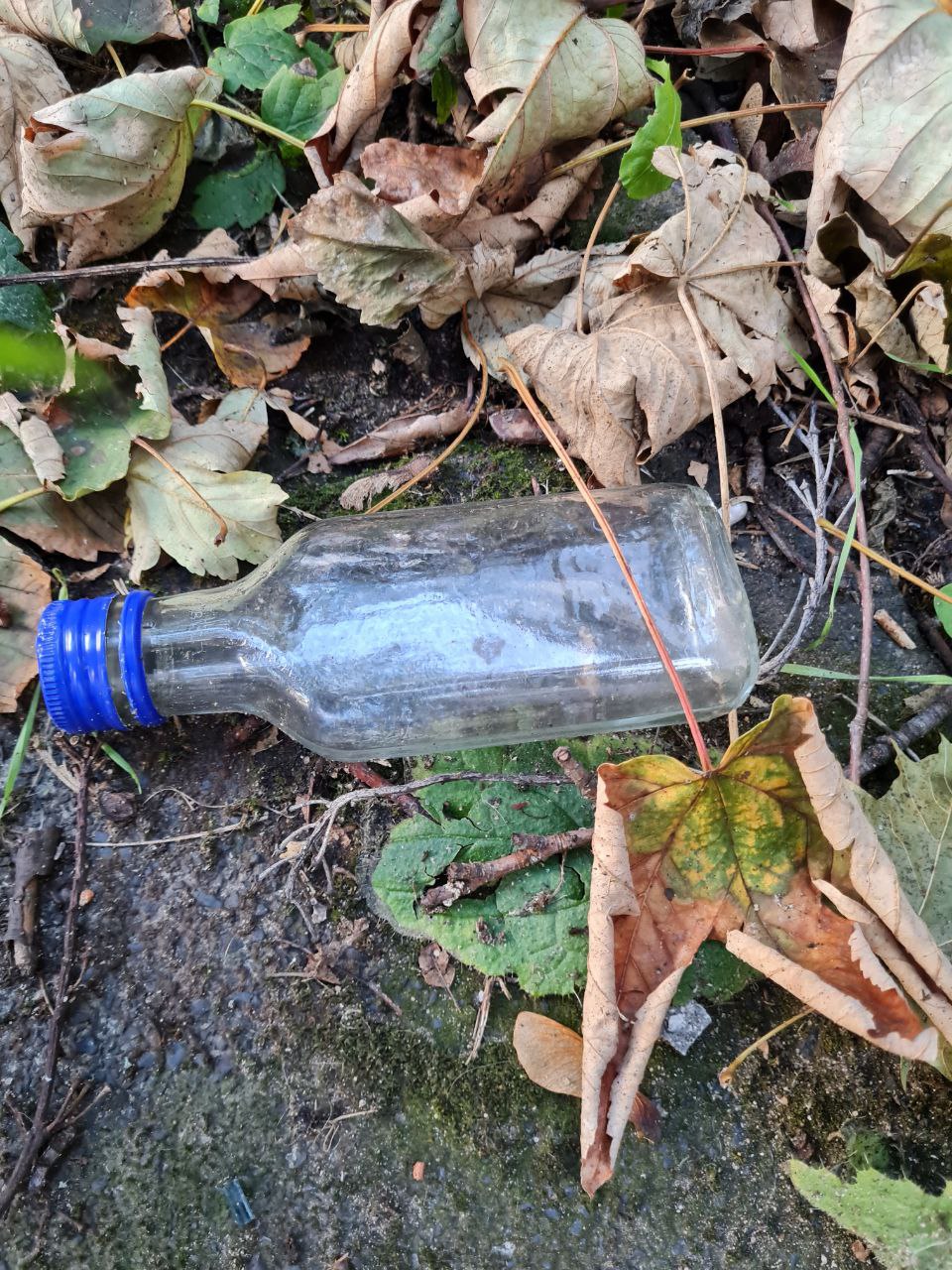}
    \includegraphics[width=0.24\linewidth]{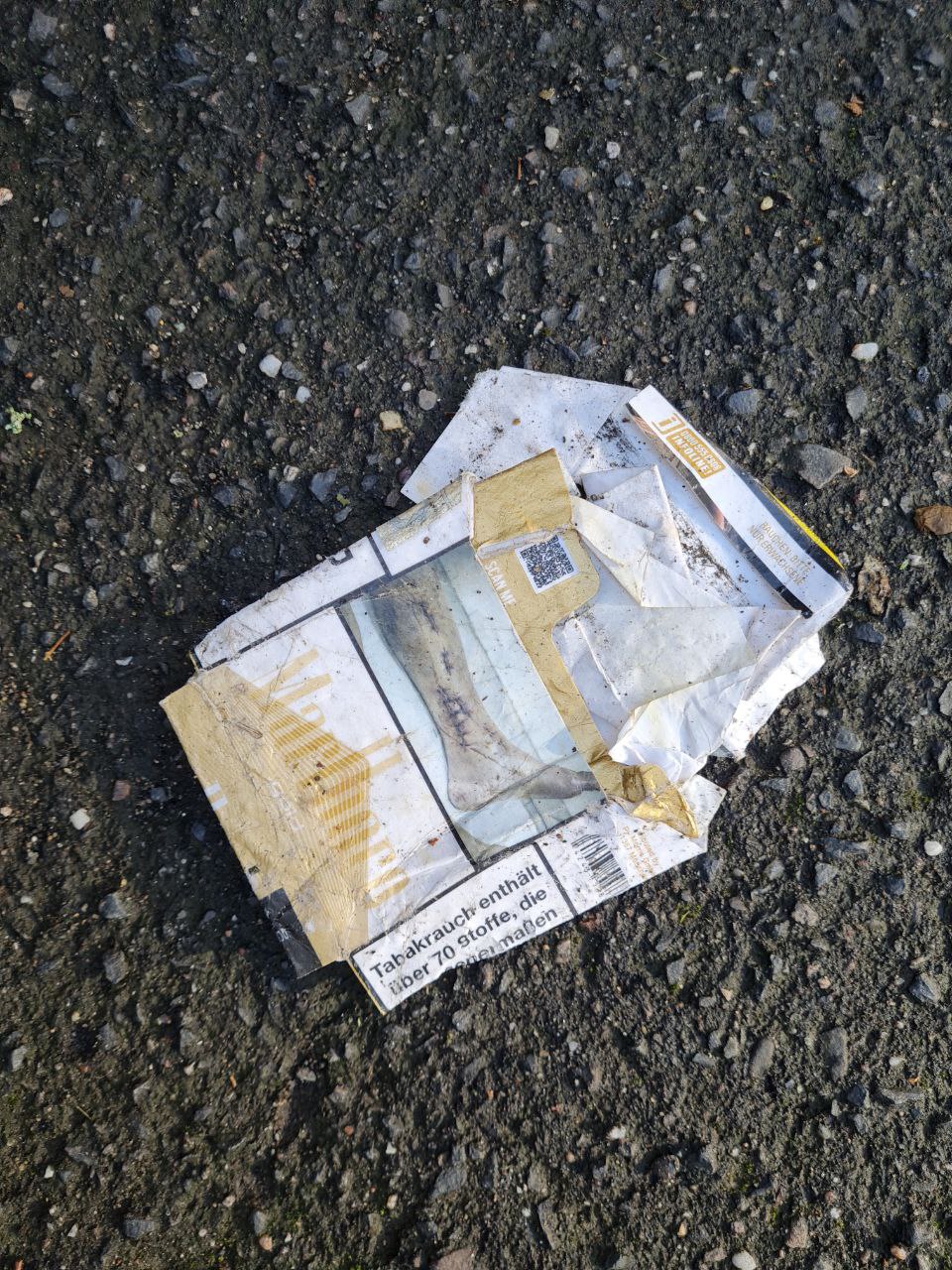}
    \caption{Exemplary trash from public spaces.}
    \label{fig:trash}
\end{figure}


\definecolor{myGray}{RGB}{120,140,140}
\definecolor{myGrayParkbucht}{RGB}{160,160,160}
\definecolor{myGreen}{RGB}{0,200,71}
\definecolor{myBlue}{RGB}{0,71,255}
\definecolor{myRed}{RGB}{255,0,0}
\definecolor{myWhite}{RGB}{255,255,255}
\definecolor{myCyan}{RGB}{200,172,255}
\definecolor{myOrange}{RGB}{255,127,0}

\begin{figure}
    \centering
    \begin{tikzpicture}[%
        every node/.style={
          font=\scriptsize,
        },
    ]
        \pgfmathsetseed{7}
        \pgfmathsetmacro{\pathwaywidth}{3}
        \pgfmathsetmacro{\wayside}{1}
        \fill[myGreen] (0,0) rectangle (11,\wayside);
        \fill[myGray] (0,\wayside) rectangle (11,\wayside+\pathwaywidth);
        \fill[myGrayParkbucht] (2,\wayside) rectangle ++ (0.1,1.5) node[xshift=-0.3cm, yshift=-0.9cm,anchor=west,midway,rotate=90] {\textcolor{black}{Curbside}};
        \fill[myGreen] (0,\pathwaywidth+\wayside) rectangle (11,\pathwaywidth+2*\wayside);
        \fill[myOrange] [rounded corners=0.1cm] (0.2,1) rectangle ++(1.2,0.75) node [midway] {\textcolor{black}{Robot}};
        \draw[decorate,decoration={brace,amplitude=5pt}] (0,\wayside) -- (0,\wayside+\pathwaywidth)
        node[xshift=-0.5cm, yshift=-0.4cm,anchor=west,midway,rotate=90] {Pathway};
        \draw[decorate,decoration={brace,amplitude=5pt}] (0,0) -- (0,\wayside)
        node[xshift=-0.5cm, yshift=-.7cm,rotate=90,anchor=west, midway] {Wayside (\wayside\,m)};
        \fill[myGrayParkbucht] [rounded corners=0.1cm] (6,\pathwaywidth+\wayside) rectangle ++ (2,+1) node [midway] {\textcolor{black}{Parking lot}};
        \fill[myGrayParkbucht] [rounded corners=0.1cm] (8,\pathwaywidth+\wayside) rectangle ++ (2,+1);
        \fill[myCyan] [rounded corners=0.1cm] (8.1,\pathwaywidth+\wayside+0.1) rectangle ++ (+1.7,+0.7) node [midway] {\textcolor{black}{Car}}; 
        \foreach \i in {1,...,20}{
          \pgfmathsetmacro{\x}{2+rnd*6}
          \pgfmathsetmacro{\y}{rnd*4.25}
          \pgfmathsetmacro{\r}{200}
          \pgfmathsetmacro{\g}{100}
          \pgfmathsetmacro{\b}{0}
          \definecolor{myColor}{RGB}{\r,\g,\b}
          \fill[myColor] [rounded corners=0.1cm] (\x,\y) rectangle (\x+0.1+rnd*0.2,\y+0.1+rnd*0.2);
        }
        \draw[pattern=north west lines, pattern color=red] (4, \wayside) rectangle (4.2,\wayside+\pathwaywidth/2);
        \draw[decorate,decoration={brace,amplitude=5pt}] (4, \wayside) -- (4,\wayside+\pathwaywidth/2) node[xshift=-0.5cm, yshift=-0.9cm,anchor=west,midway,rotate=90] {Cycle barrier};
        \draw[pattern=north west lines, pattern color=red] (5, \wayside+\pathwaywidth/2) rectangle (5.2,\wayside+\pathwaywidth);
        \fill[myBlue] [rounded corners=0.1cm] (7,\wayside) rectangle ++ (1,+0.4) node [midway] {\textcolor{black}{Bank}};
    \end{tikzpicture}
    \vspace*{-2ex}
    \caption{Scenario layout sketch.}
    \label{fig:scenario}
\end{figure}
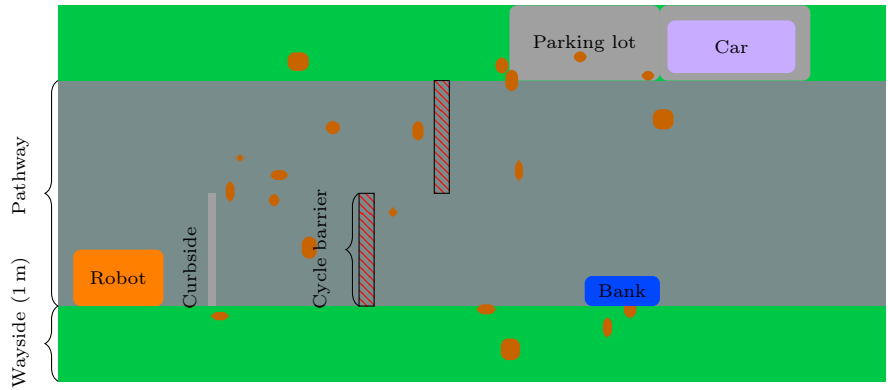

In Figure~\ref{fig:scenario}, we depict a potential setup of a park cleaning benchmark. A pathway containing various types of pollution, and the wayside (up to 1m distance from the pathway) are subject to being cleaned. The benchmark aims to take place in a real park environment.
The contamination is caused by e.g. glass, plastic cups, paper, cigarette butts, etc. (see Figure~\ref{fig:trash}) and differs e.g. by size, material, moisture, appearance, visibility. Bonus points can be awarded for removing hard-to-reach litter, e.g. under park benches or behind obstacles. 
Various obstacles (like a curbside, cycle barriers, banks and parking cars) benchmark the handling of obstacles and the cleaning capabilities in hard-to-reach areas (below bank or car). We found this setting to reflect various technical challenges and being highly practical at the same time.
The benchmark scoring considers the amount of collected items (in weight or amount) or the time required to achieve the cleaning task.


\paragraph{Technical requirements}

The benchmark is designed to be platform-independent, such that existing platforms can be used. For example, it is possible to take part in the benchmark with unmanned aerial vehicles, unmanned ground vehicles, walking robots or even combinations of several platforms. The technical requirements are checked in the form of a technical acceptance (inspection) at the start of the competition. Participation in the rest of the competition is only possible if all technical requirements are met. The technical requirements are defined as follows:
\begin{itemize}
    \item Weight $\leq$ 150\,kg
    \item Size $\leq$ 1.2~m $\times$ 0.75~m $\times$ 1~m (L$\times$W$\times$H)
    \item Well visible and reachable emergency button
    \item Obstacle avoidance (avoiding people, animals, pathway users and other obstacles) 
\end{itemize}

These requirements are inspired by non-autonomous cleaning machines for public spaces and mobility scooters.
Robots being benchmarked should follow size and weight limitations. Further, it is required that benchmarked robots can move on the pathway and potentially the wayside without damaging them. At all time, participating robots should be able to be evacuated from the benchmarking field promptly.

\section{Conclusion}

In this paper, we discuss the application of cleaning robots for public spaces. Stakeholder interviews were conducted to identify application requirements. Based on that, we derived requirements for basic technologies. 
The interviews and requirements have been used as the foundation for the design of a practical benchmark for the comparison of cleaning robots for public spaces.
These findings underscore the diverse application potential of robotic cleaning systems in public spaces, addressing practical cleaning needs and technical challenges.



\begin{credits}
\subsubsection{\ackname} 
This work has been funded by the German Ministry of Education and Research
(BMBF), grant nos. 16SV8680, 16SV8681, 16SV8683 project: Transferzentrum Roboter im Alltag (RimA).
\end{credits}

\bibliographystyle{splncs04}
\bibliography{bibliography}

\end{document}